\pdfoutput=1
\documentclass[10pt]{svproc}
\usepackage{url}

\usepackage{color}% The package I import.
\usepackage{graphicx} % The package I import.
\usepackage{amsmath} % The package I import.
\usepackage{multicol} % The package I import.
\usepackage{footmisc} % The package I import.

\usepackage{hyperref} % For ArXiv
\usepackage{cleveref} % For ArXiv

\begin{document}
\mainmatter              % start of a contribution
\title{Gradient Boost with Convolution Neural Network for Stock Forecast}
\titlerunning{Gradient Boost}  % abbreviated title (for running head)
\author{Jialin~Liu\inst{1} \and Chih-Min Lin\inst{2}
\and Fei Chao\inst{1}
}
%
%\authorrunning{Ivar Ekeland et al.} % abbreviated author list (for running head)
%\tocauthor{Ivar Ekeland, Roger Temam, Jeffrey Dean, David Grove, Craig Chambers, Kim B. Bruce, and Elisa Bertino}
%

\institute{Cognitive Science Department Xiamen University Fujian, P. R. China 361005 \\
\email{31520171153232@stu.xmu.edu.cn}, \email{fchao@xmu.edu.cn}
\and
Department of Electrical Engineering Yuan Ze University \\
Chung-Li, Tao-Yuan 320, Taiwan \\
\email{cml@saturn.yzu.edu.tw}}

\maketitle              % typeset the title of the contribution

\begin{abstract}
   Market economy closely connects aspects to all walks of life. The stock forecast is one of task among studies on the market economy. However, information on markets economy contains a lot of noise and uncertainties, which lead economy forecasting to become a challenging task. Ensemble learning and deep learning are the most methods to solve the stock forecast task. In this paper, we present a model combining the advantages of two methods to forecast the change of stock price. The proposed method combines CNN and GBoost. The experimental results on six market indexes show that the proposed method has better performance against current popular methods.
% We would like to encourage you to list your keywords within
% the abstract section using the \keywords{...} command.
\keywords{ensemble learning, deep learning, stock forecast}
\end{abstract}
\section{Introduction}
The stock price of a company is an important criterion for measuring the actual value of the company. In the stock market, well decision depends on well forecast.
%In the stock market, if an investor wants to be profitable, the stock must be bought and sold at the right time. The price of stock is changing all the time. Therefore, we desire a method to forecasting future price of stock accurately so as to make a better decisions.
Due to the development of computational technology and intelligence technologies. New tools are recently developed to process information on stock forecast.
%As a new kind of investment analysis method which combines modem financial data quantitative trading and mathematics is playing an active and significant role in financial markets all over the world \cite{li2016quantitative}.
The analysis of financial market movements has been widely studied in the fields of finance, engineering and mathematics in the last decades \cite{porshnev2013machine}. Using intelligent technologies on stock prediction has widely spread in recent years.

In the past, the most commonly algorithms for stock forecast in the past are artificial neural network (ANN) and support vector machine (SVM or SVR).
In contract to ANN, SVM is a statistical learning method that is widely used in pattern recognition tasks. In 2003, Kim et al. predicted stock price by using a SVR model and proved that the prediction precision of support vector regression model was better than the back propagation (BP) neural network prediction model and case-based reasoning (CBR) \cite{kim2004toward}. In 2006, Xu et al. came up with a revised Least squares (LS)-SVM model and forecasted Nasdaq Index movement, and model brought satisfactory results \cite{rui2006nonlinear}.

%A research about ANN in business shows that the number of articles published by the ANN for financial use has slowly increased year by year \cite{tkavc2016artificial}.
In the past several years, there are many business applications, in which the technology of ANN was used. It may be due to the non-linear approximation ability of ANN, and it is frequently used in combination with other methods. Martinex et al. proposed that the neural network to solve the financial forecasting problem use mostly a back propagation algorithm to optimize a multi-layer forward neural network (MLP) with high performance \cite{martinez2009artificial}.
%For example, a fusion of machine learning techniques containing ANN and SVR for market analysis is presented by \cite{patel2015predicting} gets a well result.

In recent years, ensemble learning and deep learning have developed quickly in a lot of fields \cite{chen2016xgboost,lawrence1997using,diao2013feature,nassirtoussi2015text,he2016deep}. These two methods have their own advantages and disadvantages in solving with stock data. In general, there are two kinds of different views in stock prediction. (i) To obtain the enough information, financial analysis methods must be used to obtain high quality information. In this case, market economy data involves the indices with different characters, which are suitable to be handed by ensemble learning. However, simplifies computation of analysis methods would causes loss of information. (ii) Only use the stock price history to get information. It is possible that all the information is available from the historical behavior of a financial asset as a time series. The time series of stock prices involves enough information and is suitable to be handed by deep learning. However, the time series of stock prices also involves a lot of noise and uncertainties.
%Since, market economy data is usually organized as time series. Deep learning has advantages to solve this kind of sequence data, which has local correlation different amongst dimensions.
%However, these two methods have their own disadvantages. (i) Financial analysis methods usually involve some prior hypothesis, which simplifies computation but causes loss of information. (ii) Only using the stock price history to get the future price is also a difficult task. A lot of noise and uncertainties exit in the stock's historical price time series.
In order to improve forecast accuracy we desire to use technique indices and retain the sequence structure of stock data to make them complementary to each other.
Thus, a model combines the advantages of both ensemble learning and deep learning is the objective of this paper. This paper proposes a model combining the ensemble techniques of extreme gradient boost (XGBoost)~\cite{chen2016xgboost} and 1D convolution neural network (CNN), called as CGBoost, to obtain a better performance.

In addition, we use a sparse autoencodes (SAEs)~\cite{hinton2006reducing,bengio2007greedy} to process data to reduce noise in the stock price time series, the training implemented by encoding and decoding data and reducing the loss in each iteration. If we only and use the original data to train CGBoost without this process, we will get only a highly overfitting result. Then we also try to training one model on several different market indexes, so as to test whether the propose model can unify data from different market indexes to improve overall performance.

The remainder of this paper contains three sections. Section \ref{methology} draws the details of each technique used in this work and how to combine all those techniques as a complete system. Section \ref{experiment} describes data resources, evaluation and other details about the experiment. The results and analysis of the experiment is also in this section. Finally, section \ref{conclusion} draws conclusion and future work.

%------------------------------------------------------------------------
\section{Methology}\label{methology}

\begin{figure*}[t]
  \centering
  \includegraphics[width=\linewidth]{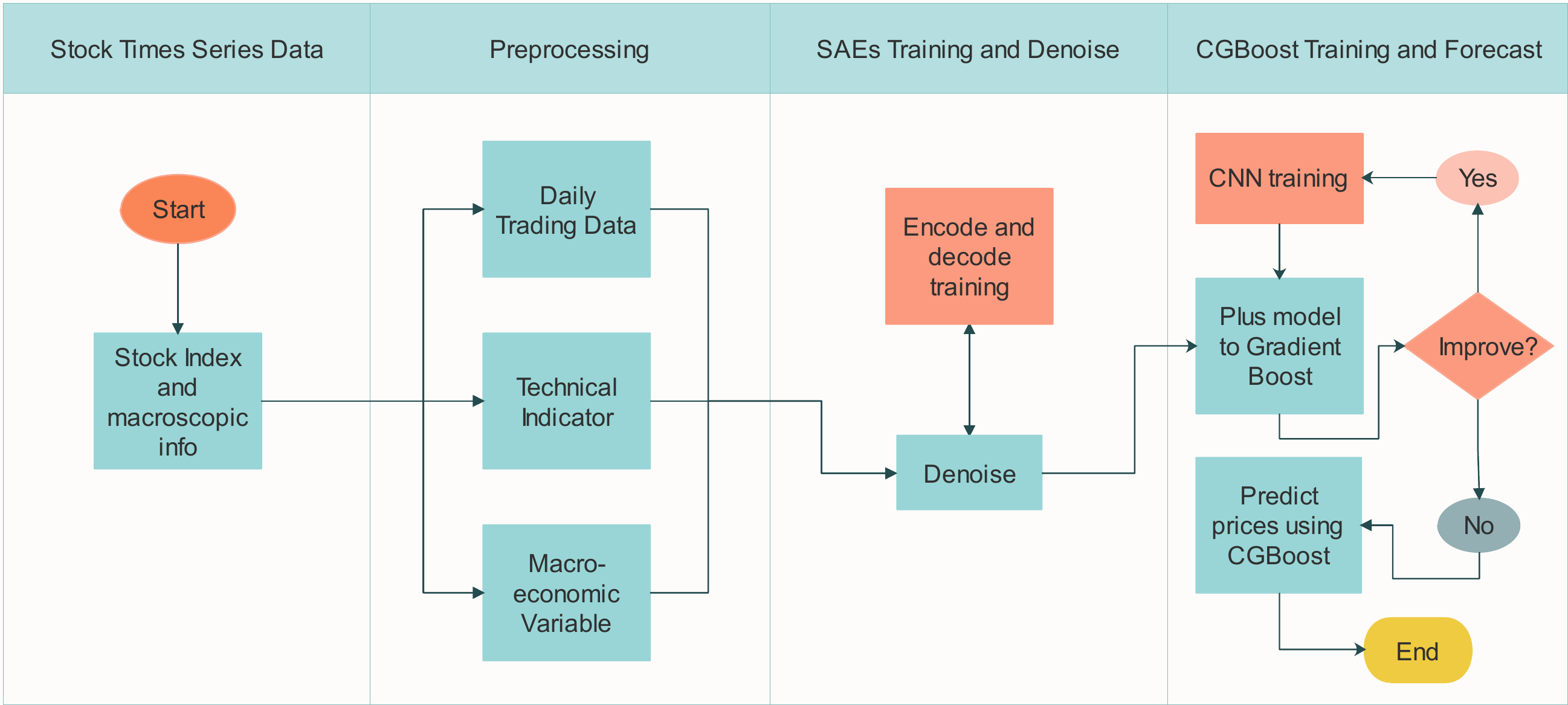}
  \caption{Stock prediction system contains sparse autoencodes and convolution gradient boost.}
  \label{framesystem}
\end{figure*}

%------------------------------------------------------------------------
%\subsection{Full Frame}
In order to generate the deep and invariant features for one-step-ahead stock price forecast, this paper presents a gradient boosting framework with a deep learning for financial time series. The framework uses a deep (CNN) and width (GBoost) learning-based predicting scheme that integrates the architecture of CNN and GBoost. The flow chart of this framework shown in Fig. \ref{framesystem}, involves three stages: (1) data preprocessing, the clipping and normalizing transform, which are applied to rescale the stock price time series to some scale; (2) adopting of the SAEs, which has a deep architecture trained in an unsupervised manner, combined with 1D CNN; and (3) GBoost, it has 1D CNN to generate the one-step-ahead prediction. Since the first step is related to data descriptions, details the first step are introduced in the Section \ref{experiment}. The rest steps are detailed as follows.

%-------------------------------------------------------------------------
\subsection{SAEs Training and Denoise}

SAEs is a type of deep learning model to reduce dimension and noise of data~\cite{hinton2006reducing,bengio2007greedy}. Since manually adding category tags to data is a very cumbersome process, the machine must learn part of the important features in the sample. By imposing some restrictions on the hidden layer, SAEs can better express the characteristics of the sample in a harsh environment. SAEs has this limitation on the sparseness of the hidden layer.

The sparsity is represented as the activated states of neuron. If the sigmoid function is used as the activation function, and the neuron output value is 0, this situation is regarded as a suppression. The sparsity limit ensures that most of the neuron output is 0 and the state is suppressed. Then, the functions can approximate,

\begin{equation}
\hat{\rho}=\rho,\ \hat{\rho}=\frac{1}{m}\sum^m_{i=1}\left[a_j(x^{(i)})\right]
\end{equation}
where $a_j$ denotes the activation of the hidden neuron $j$; $\hat{\rho}$ is the average of the activation; and $\rho$ is a sparsity parameter, usually it is a small value close to 0 (such as $\rho = 0.05$).

In order to achieve this limitation, an additional penalty factor is added to our optimization objective function, which can punish those $\hat\rho_j$ has significantly different conditions with $\rho$ in hidden layers, it is given by:

\begin{equation}
\sum_{j=1}^{s_2} {\rm KL}(\rho || \hat\rho_j)=
\sum_{j=1}^{s_2} \rho \log \frac{\rho}{\hat\rho_j} + (1-\rho) \log \frac{1-\rho}{ 1-\hat\rho_j},
\end{equation}
where $s_2$ is the number of hidden neurons in the hidden layer; and the index $j$ in turn represents each neuron in the hidden layer. Then, the overall loss function is expressed as:
\begin{equation}
J_{\rm sparse}(W,b) = J(W,b) + \beta \sum_{j=1}^{s_2} {\rm KL}(\rho || \hat\rho_j),
\end{equation}
where $J(W,b)$ represents the reconstruction loss; $\beta$ controls the weight of the sparsity penalty factor, $W$ and $b$ are weight and bias of neural network, respectively.

Finally, we apply stochastic gradient descend to optimize $W$ and $b$. In order to minimize $J_{\rm sparse}(W,b)$. After training SAEs, ${\rm a}(x^{(i)})={a_j(x^{(i)})}_j$ is used as the feature of sample $\{x^{(i)},y^{(i)}\}$.

\subsection{CGBoost Training and Forecast}

The gradient boost algorithm is an ensemble learning technology. The algorithm generates a prediction model by integrating weak prediction models, such as decision trees. It builds the model in a step-by-step manner like other gradient methods and promotes them by allowing the use of any differentiable loss function.

In the experiments, rather than using original GBoost, we apply the training way in XGBoost~\cite{chen2016xgboost}. Different from the traditional GBoost method, only the first derivative information is used. XGBoost performs the second-order Taylor expansion on the loss function, and adds the regular term to the objective function to balance the decline of the objective function, so as to avoid overfitting. The objective function of the based learner is given by:
%\begin{equation}
%\begin{aligned}
%Obj^{(t)}&=\sum^n_{i=1}l(y_i,y_i^{(t-1)}+f_t(x_i))+\Omega(f_t)+constant\\
%&\approx\sum^n_{i=1}\left[l(y_i,y^{(t-1)}_i)+\partial_{y^{(t-1)}_i}l(y_i,y^{(t-1)}_i)f_t(x_i)\right.\\
%&\left.+\frac{1}{2}\partial^2_{y^{(t-1)}_i}l(y_i,y^{(t-1)}_i)f^2_t(x_i)\right]+\Omega(f_t)+constant.
%\end{aligned}
%\end{equation}

\begin{equation}
Obj^{(t)}\approx\sum^n_{i=1}\left[g_if_t(x_i)+\frac{1}{2}h_if^2_t(x_i)\right]+\Omega(f_t),
\end{equation}
where $\Omega(f_t)$ is the L2 regularization $\sum_l \|W_l\|^2$, due to all based estimators are CNNs, $W_l$ denotes the weights of $l$ layer, $g_i=\partial l(y_i,y^{(t-1)_i})/\partial y^{(t-1)}_i$ and $h_i=\partial^2 l(y_i,y^{(t-1)_i})/\partial {y^{(t-1)}_i}^2$. Because the goal is to predict the real price of stock the loss function can be square loss function. Then, the form is given by:
\begin{equation}
Obj^{(t)}\approx\sum^n_{i=1}\left[2(y_i^{(t-1)}-y_i)f_t(x_i)+f^2_t(x_i)\right]+\Omega(f_t),\ t\neq1
\end{equation}
when all based estimators are obtained, the forecast result of $x_i$ is calculated by $F(x_i)=\sum^T_{t=1}f_t(x_i)$, where $T$ denotes the number of based estimators.

\begin{figure*}[t]
  \centering
  \includegraphics[width=\linewidth]{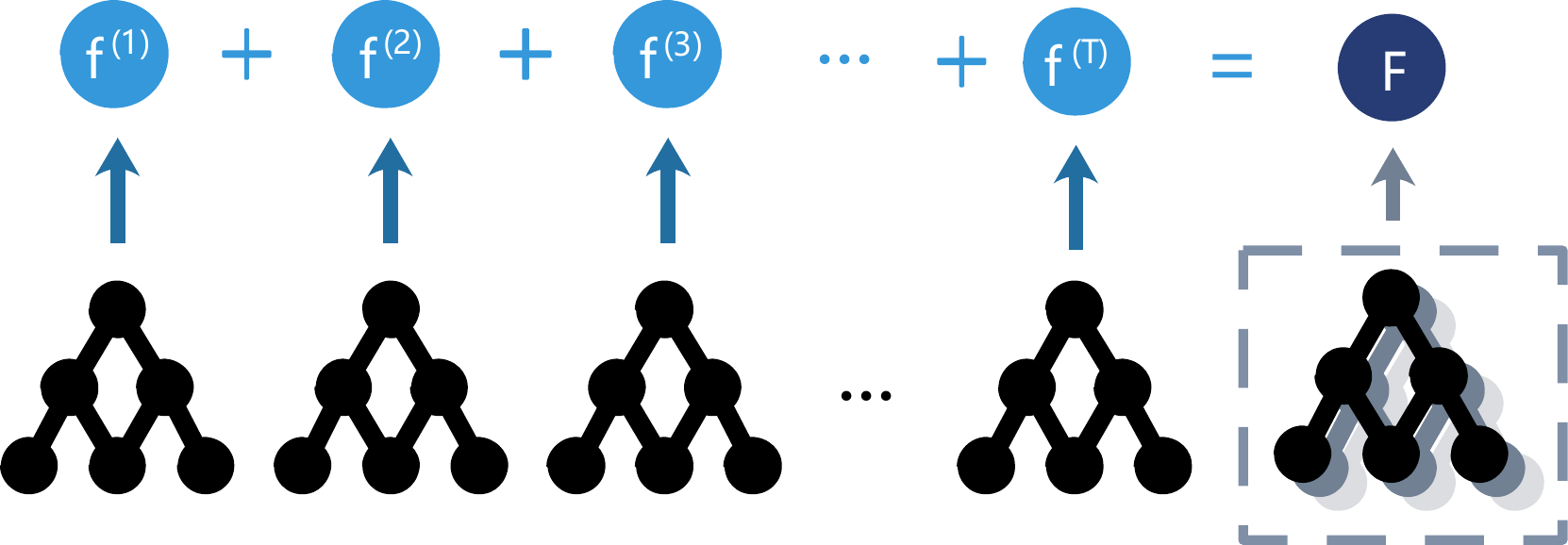}
  \caption{Ensemble model with neural network based learners.}
  \label{GBoost}
\end{figure*}

%If we want to train a two classify evaluator, which is used to predict the up and down of stock time series, then the forms of $g_i$ and $h_i$ is given by,
%
%\begin{equation}
%\begin{aligned}
%g_i&=\partial_{y_i^{(t-1)}}l(y_i,y^{(t-1)}_i)\\
%&=-y_i(1-\frac{1}{1+e^{-y_i^{(t-1)}}})+(1-y_i)\frac{1}{1+e^{-y_i^{(t-1)}}}
%\end{aligned}
%\end{equation}
%\begin{equation}
%h_i=\partial^2_{y_i^{(t-1)}}l(y_i,y^{(t-1)}_i)=\frac{e^{-y^{(t-1)}_i}}{\left(1+e^{-y_i^{(t-1)}}\right)^2}
%\end{equation}
%Then loss function is given by,
%\begin{equation}
%\begin{aligned}
%&Obj^{(t)}\approx\\
%&\sum^n_{i=1}\Bigg[\left(-y_i(1-\frac{1}{1+e^{-y_i^{(t-1)}}})+(1-y_i)\frac{1}{1+e^{-y_i^{(t-1)}}}\right)f_t(x_i)\\
%&+\frac{1}{2}\frac{e^{-y^{(t-1)}_i}}{\left(1+e^{-y_i^{(t-1)}}\right)^2}f^2_t(x_i)\Bigg]+\Omega(f_t)\ ,\ t\neq1
%\end{aligned}
%\end{equation}
%
%In experiment model is used to predict the next day price, so we applied the object function $(8)$. In Figure 4 it is the structure of our predictor. Because the base learners used in Boost are usually many weak learners, we use few channels in 1D resnet to make it become weak learners.

%-------------------------------------------------------------------------
\subsection{1D Residual Network}
In this paper, we use 1D residual neural network (resnet)~\cite{he2016deep}, a kind of CNN, within both SAEs and GBoost. Due to the common size of our model we use resnet to train well. However, using resnet still can accelerate training significantly~\cite{he2016deep}.

\begin{figure*}[t]
  \centering
  \includegraphics[width=\linewidth]{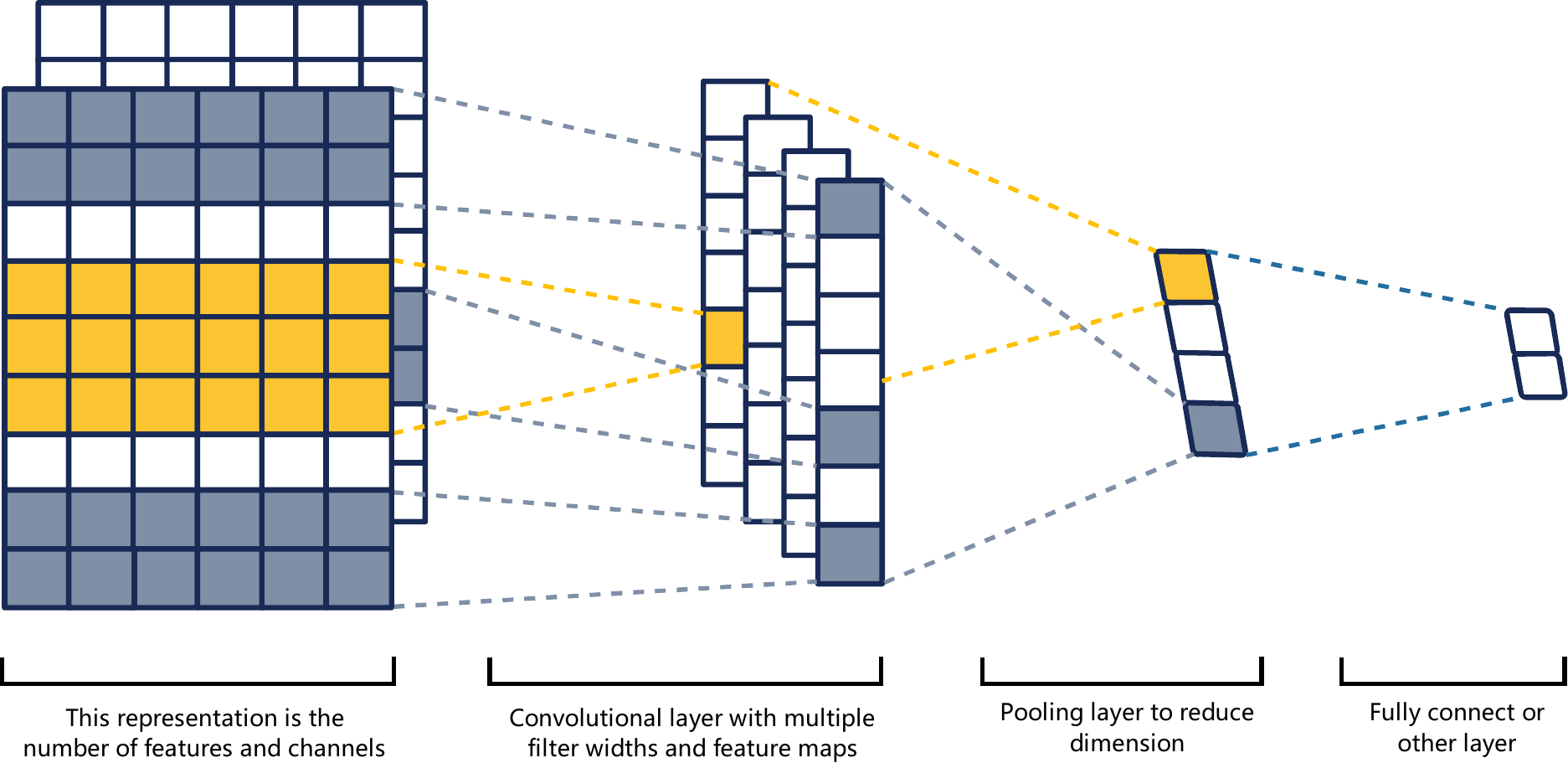}
  \caption{A standard 1D convolutional neural network diagram.}
  \label{1DCNN}
\end{figure*}

A structural diagram of a standard 1D CNN is shown in Fig. \ref{1DCNN}. Each layer receives the output from the previous layer and outputs abstract features. In the training, gradient is back propagated from the output of last layer. The level number of network is larger than a certain number, the gradient vanishing will occur, so as to make deep network to be trained difficultly.

The resnet applies the idea of the ``shortcut connections'', the idea of cross-layer linking to improve it, in order to prevent the gradient vanishing. The input $x$ is directly passed to the output as the initial result, and the output result is $H(x)=F(x)+x$. If $F(x)=0$, $H(x)$ will become the identity map, $H(x)=x$. As the network deepening, it still retain the much shallower tunnel. Therefore, the gradient does not decrease as the network deepening.

%------------------------------------------------------------------------
\section{Experiment}\label{experiment}
The experiments are designed to answer two questions: (1) Can the model combine ensemble learning and deep learning produce more accurate predictions than single deep learning model? (2) Is proposed model able to fit the data from different indices and still improve the performance?

The proposed model compares with the accuracy of WSAEs-LSTM \cite{bao2017deep}, which applied the deep learning model to forecast the stock price series, so as to answer the first question. Following \cite{bao2017deep}, we chose ``CSI 300'', ``DJIA'', ``Hang Seng'', ``Nifty 50'', ``Nikkei 225'' and ``S\&P500'' indices as the predict goals. We conducted experiments training one model for each index and training one model for all indexes, and CGBoost and CGBoost6 were applied to denote them respectively. Their results can answer the second question.

It is different from Fig. \label{framesystem} we use a fixed number of base models in CGBoost. The reason is that we train the model on training data and validation data after adjusting hyper-parameter. Thus not validation data can be used to test whether the model is improved or not. Besides, the model is used to predict stock price indirectly by predicting the change rate of price. Based on our experience, this way can get better results.

%-------------------------------------------------------------------------
\subsection{Data descriptions}
The data used in this experiment is detailed as follows.

\textbf{Data resource.} We use the data provided by \cite{bao2017deep}, which was sample from CSMAR$\footnote{\url{http://www.gtarsc.com}}$ and WIND$\footnote{\url{http://www.wind.com.cn}}$. The sample is from $1^{\rm st}$ Jul. 2008 to $30^{\rm th}$ Sep.2016.

\begin{table*}
\begin{center}
\begin{tabular}{c|l}
\hline
Name& \multicolumn{1}{c}{Definition} \\
\hline
MACD & Moving Average Convergence \\
\hline
WVAD & Williams's Variable Accumulation/Distribution \\
\hline
ATR & Average true range \\
\hline
EMA20 & 20 day Exponential Moving Average \\
\hline
BOLL & Bollinger Band MID \\
\hline
MA5/MA10 & 5/10 day Moving Average \\
\hline
MTM6/MTM12 & 6/12 month Momentum \\
\hline
SMI & Stochastic Momentum Index \\
\hline
ROC & Price rate of change \\
\hline
CCI & Commodity channel index \\
\hline
\end{tabular}
\end{center}
\caption{The techniques indices and their definition is described in this table.}
\label{indexfeat}
\end{table*}

\textbf{Data features.} Three types of feature are chose in our experiment. Following the previous literature the first type of feature includes OHLC variables, which is the price variables (Open, High, Low, and Close price). The second kind of feature is the technical indicators of each index. Each of them is described in Table \ref{indexfeat}. The final part of inputs is the macroeconomic variable. It is related to stock price. We chose the Interbank Offered Rate and US dollar Index to our system.

\textbf{Data divide.} Refer to the rule of stock, we cannot use the data from future. Thus we use the first two year as the training set, next three months as the validating data and last three months as the test set. It is divided into four steps to obtain a one-year prediction result for testing. We divide prediction into six years to evaluate accuracy.% (see Table 2).

%\renewcommand\arraystretch{1.}
%\begin{table}
%\begin{center}
%\begin{tabular}{c|l}
%\hline
%Year & \multicolumn{1}{c}{Time Interval} \\
%\hline
%Year1 & 2010.10.01$-$2011.09.30 \\
%Year2 & 2011.10.01$-$2012.09.30 \\
%Year3 & 2012.10.01$-$2013.09.30 \\
%Year4 & 2013.10.01$-$2014.09.30 \\
%Year5 & 2014.10.01$-$2015.09.30 \\
%Year6 & 2015.10.01$-$2016.09.30 \\
%\hline
%\end{tabular}
%\end{center}
%\caption{The prediction time interval of each year.}
%\end{table}

\subsection{Evaluate}
Following \cite{guo2014feature,Hsieh2011forecasting,altay2005stock,emenike2010forecasting}, the results were evaluated by ``MAPE'', ``Theil U'' and ``linear correlation between prediction and real prices'' (use R to denote). These indicators are denoted as follows:

\begin{equation}
\begin{aligned}
\mathrm{MAPE}=\frac{1}{N}\sum^N_{t=1}\left|\frac{y_t-y_t^*}{y_t}\right|
\end{aligned}
\end{equation}

\begin{equation}
\begin{aligned}
\mathrm{R}=\frac{\sum^N_{t=1}(y_t-\overline{y_t})(y^*_t-\overline{y^*_t})}{\sqrt{\sum^N_{t=1}(y_t-\overline{y_t})^2\sum^N_{t=1}(y^*_t-\overline{y^*_t})^2}}
\end{aligned}
\end{equation}

\begin{equation}
\begin{aligned}
\mathrm{Theil\ U}=\frac{\sqrt{\frac{1}{N}\sum^N_{t=1}(y_t-y_t^*)^2}}{\sqrt{\frac{1}{N}\sum^N_{t=1}(y_t)^2}+\sqrt{\frac{1}{N}\sum^N_{t=1}(y^*_t)^2}}
\end{aligned}
\end{equation}
where $y^*_t$ is the forecast of model and $y_t$ is the actual price on time $t$. N is the number of prediction, in our experiment it is the number of days open in a year. R is different from MAPE and Theil U, if R is larger, the predicting price is similar to the actual value.

%\subsection{Based experiment}
%We use a lower weight decay in basic convolution neural network in order to get high prediction accuracy. In experiment we get the highest accuracy in training set, the mean of MAPE in each index is lower than 0.005, Theil U is lower than 0.003 and linear correlation is closed to 1. But in validating and testing data the mean of MAPE is higher than 0.023 in each index, Theil U is higher than 0.018 and linear correlation is lower than 0.6.
%
%We can see this is a result with high overfit. The loss in training data is much higher than validating and testing data. We think the lower weight decay is probably reason. So in later experiment we adjusted the weight decay to see the result.
%
%\subsection{Weight decay effect to performance}
%
%Then we increase the weight decay in basic convolution neural network in order to reduce overfitting. The experiment result is in Table 3-8. Within each table, our three accuracy indicators are demonstrates to three panels, each one is one of accuracy indicators. We report the six yearly results and the average measure result over the six years in a table for one stock index. The results of different index are in the different tables.

\renewcommand\arraystretch{1.}
\begin{table}
\begin{center}
\resizebox{1.\hsize}{!}{
\begin{tabular}{c|c|c|c|c|c|c|c|c|c|c|c|c|c|c}
\hline
 & \multicolumn{7}{c|}{CSI 300 index} & \multicolumn{7}{c}{DJIA index}\\
\hline
Year & Year 1 & Year 2 & Year 3 & Year 4 & Year 5 & Year 6 & Average & Year 1 & Year 2 & Year 3 & Year 4 & Year 5 & Year 6 & Average \\
\hline
\multicolumn{15}{c}{Panel A.MAPE} \\
\hline
CGBoost6 & \textbf{0.011} & \textbf{0.011} & \textbf{0.010} & \textbf{0.008} & \textbf{0.018} & \textbf{0.011} & \textbf{0.011} & \textbf{0.008} & \textbf{0.008} & \textbf{0.005} & \textbf{0.005} & \textbf{0.007} & \textbf{0.007} & \textbf{0.007} \\
\hline
CGBoost & 0.015 & 0.012 & 0.014 & 0.010 & 0.024 & 0.014 & 0.015 & 0.011 & 0.010 & 0.008 & 0.008 & 0.011 & 0.011 & 0.010 \\
\hline
WSAEs-LSTM & 0.025 & 0.014 & 0.016 & 0.011 & 0.033 & 0.016 & 0.019 & 0.016 & 0.013 & 0.009 & 0.008 & 0.008 & 0.010 & 0.011 \\
\hline
\multicolumn{15}{c}{Panel B.Correlation coefficient} \\
\hline
CGBoost6 & \textbf{0.974} & \textbf{0.971} & \textbf{0.978} & \textbf{0.978} & \textbf{0.992} & \textbf{0.975} & \textbf{0.978} & \textbf{0.972} & \textbf{0.977} & \textbf{0.994} & \textbf{0.983} & \textbf{0.962} & \textbf{0.976} & \textbf{0.977} \\
\hline
CGBoost & 0.953 & 0.963 & 0.965 & 0.965 & 0.988 & 0.967 & 0.967 & 0.949 & 0.964 & 0.987 & 0.956 & 0.919 & 0.939 & 0.952 \\
\hline
WSAEs-LSTM & 0.861 & 0.959 & 0.955 & 0.957 & 0.975 & 0.957 & 0.944 & 0.922 & 0.928 & 0.984 & 0.952 & 0.953 & 0.952 & 0.949 \\
\hline
\multicolumn{15}{c}{Panel C.Theil U} \\
\hline
CGBoost6 & \textbf{0.007} & \textbf{0.007} & \textbf{0.007} & \textbf{0.005} & \textbf{0.013} & \textbf{0.008} & \textbf{0.008} & \textbf{0.006} & \textbf{0.005} & \textbf{0.003} & \textbf{0.003} & \textbf{0.005} & \textbf{0.004} & \textbf{0.004} \\
\hline
CGBoost & 0.010 & 0.008 & 0.009 & 0.006 & 0.016 & 0.010 & 0.010 & 0.008 & 0.007 & 0.005 & 0.005 & 0.007 & 0.007 & 0.007 \\
\hline
WSAEs-LSTM & 0.017 & 0.009 & 0.011 & 0.007 & 0.023 & 0.011 & 0.013 & 0.010 & 0.009 & 0.006 & 0.005 & 0.005 & 0.006 & 0.007 \\
\hline
\end{tabular}
}
\end{center}
\caption{The prediction accuracy in CSI 300 and DJIA indices.}
\label{CSI300_DJIA}
\end{table}

\begin{table}
\begin{center}
\resizebox{1.\hsize}{!}{
\begin{tabular}{c|c|c|c|c|c|c|c|c|c|c|c|c|c|c}
\hline
 & \multicolumn{7}{c|}{HangSeng Index} & \multicolumn{7}{c}{Nifty 50 index}\\
\hline
Year & Year 1 & Year 2 & Year 3 & Year 4 & Year 5 & Year 6 & Average & Year 1 & Year 2 & Year 3 & Year 4 & Year 5 & Year 6 & Average \\
\hline
\multicolumn{15}{c}{Panel A.MAPE} \\
\hline
CGBoost6 & \textbf{0.010} & \textbf{0.011} & \textbf{0.007} & \textbf{0.007} & \textbf{0.010} & \textbf{0.010} & \textbf{0.009} & \textbf{0.010} & \textbf{0.009} & \textbf{0.008} & \textbf{0.006} & \textbf{0.008} & \textbf{0.007} & \textbf{0.008} \\
\hline
CGBoost & 0.014 & 0.014 & 0.011 & 0.009 & 0.016 & 0.013 & 0.013 & 0.015 & 0.013 & 0.013 & 0.016 & 0.013 & 0.011 & 0.013 \\
\hline
WSAEs-LSTM & 0.016 & 0.017 & 0.012 & 0.011 & 0.021 & 0.013 & 0.015 & 0.020 & 0.016 & 0.017 & 0.014 & 0.016 & 0.018 & 0.017 \\
\hline
\multicolumn{15}{c}{Panel B.Correlation coefficient} \\
\hline
CGBoost6 & \textbf{0.982} & \textbf{0.961} & \textbf{0.967} & \textbf{0.976} & \textbf{0.985} & \textbf{0.980} & \textbf{0.975} & \textbf{0.981} & \textbf{0.969} & 0.935 & \textbf{0.997} & 0.958 & \textbf{0.988} & \textbf{0.971} \\
\hline
CGBoost & 0.963 & 0.938 & 0.943 & 0.954 & 0.954 & 0.966 & 0.953 & 0.957 & 0.935 & 0.873 & 0.978 & 0.910 & 0.972 & 0.937 \\
\hline
WSAEs-LSTM & 0.944 & 0.924 & 0.920 & 0.927 & 0.904 & 0.968 & 0.931 & 0.895 & 0.927 & \textbf{0.992} & 0.885 & \textbf{0.974} & 0.951 & 0.937 \\
\hline
\multicolumn{15}{c}{Panel C.Theil U} \\
\hline
CGBoost6 & \textbf{0.006} & \textbf{0.007} & \textbf{0.005} & \textbf{0.004} & \textbf{0.007} & \textbf{0.006} & \textbf{0.006} & \textbf{0.006} & \textbf{0.006} & \textbf{0.006} & \textbf{0.004} & \textbf{0.005} & \textbf{0.004} & \textbf{0.005} \\
\hline
CGBoost & 0.009 & 0.009 & 0.007 & 0.006 & 0.012 & 0.008 & 0.009 & 0.010 & 0.008 & 0.008 & 0.011 & 0.008 & 0.007 & 0.009 \\
\hline
WSAEs-LSTM & 0.011 & 0.010 & 0.008 & 0.007 & 0.018 & 0.008 & 0.011 & 0.013 & 0.010 & 0.010 & 0.009 & 0.010 & 0.011 & 0.011 \\
\hline
\end{tabular}
}
\end{center}
\caption{The prediction accuracy in HangSeng and Nifty 50 indices.}
\label{HangSeng_Nifty50}
\end{table}

\begin{table}
\begin{center}
\resizebox{1.\hsize}{!}{
\begin{tabular}{c|c|c|c|c|c|c|c|c|c|c|c|c|c|c}
\hline
 & \multicolumn{7}{c|}{Nikkei 225 index} & \multicolumn{7}{c}{S\&P500 Index}\\
\hline
Year & Year 1 & Year 2 & Year 3 & Year 4 & Year 5 & Year 6 & Average & Year 1 & Year 2 & Year 3 & Year 4 & Year 5 & Year 6 & Average \\
\hline
\multicolumn{15}{c}{Panel A.MAPE} \\
\hline
CGBoost6 & \textbf{0.011} & \textbf{0.009} & \textbf{0.013} & \textbf{0.009} & \textbf{0.010} & \textbf{0.013} & \textbf{0.011} & \textbf{0.009} & \textbf{0.008} & \textbf{0.006} & \textbf{0.005} & \textbf{0.007} & \textbf{0.007} & \textbf{0.007} \\
\hline
CGBoost & 0.015 & 0.013 & 0.019 & 0.012 & 0.017 & 0.017 & 0.015 & 0.014 & 0.012 & 0.008 & 0.008 & 0.011 & 0.010 & 0.011 \\
\hline
WSAEs-LSTM & 0.024 & 0.019 & 0.019 & 0.019 & 0.018 & 0.017 & 0.019 & 0.012 & 0.014 & 0.010 & 0.008 & 0.011 & 0.010 & 0.011 \\
\hline
\multicolumn{15}{c}{Panel B.Correlation coefficient} \\
\hline
CGBoost6 & \textbf{0.966} & \textbf{0.977} & \textbf{0.994} &0.957 & \textbf{0.987} & \textbf{0.970} & \textbf{0.975} & \textbf{0.972} & \textbf{0.982} & \textbf{0.994} & \textbf{0.990} & \textbf{0.953} & \textbf{0.976} & \textbf{0.976} \\
\hline
CGBoost & 0.943 & 0.958 & 0.990 & 0.932 & 0.971 & 0.957 & 0.958 & 0.938 & 0.966 & 0.988 & 0.976 & 0.892 & 0.954 & 0.952 \\
\hline
WSAEs-LSTM & 0.878 & 0.834 & 0.665 & \textbf{0.972} & 0.774 & 0.924 & 0.841 & 0.944 & 0.944 & 0.984 & 0.973 & 0.880 & 0.953 & 0.946 \\
\hline
\multicolumn{15}{c}{Panel C.Theil U} \\
\hline
CGBoost6 & \textbf{0.008} & \textbf{0.006} & \textbf{0.009} & \textbf{0.006} & \textbf{0.007} & \textbf{0.009} & \textbf{0.007} & \textbf{0.006} & \textbf{0.005} & \textbf{0.004} & \textbf{0.003} & \textbf{0.005} & \textbf{0.005} & \textbf{0.005} \\
\hline
CGBoost & 0.010 & 0.008 & 0.013 & 0.007 & 0.011 & 0.011 & 0.010 & 0.009 & 0.008 & 0.005 & 0.005 & 0.008 & 0.006 & 0.007 \\
\hline
WSAEs-LSTM & 0.016 & 0.013 & 0.013 & 0.013 & 0.012 & 0.012 & 0.013 & 0.009 & 0.010 & 0.006 & 0.005 & 0.008 & 0.006 & 0.007 \\
\hline
\end{tabular}
}
\end{center}
\caption{The prediction accuracy in Nikkei 225 and S\&P500 indices.}
\label{Nikkei225_SP500}
\end{table}

\subsection{Results}
The proposed method has improved results significantly. As show in Table \ref{CSI300_DJIA}-\ref{Nikkei225_SP500}, both cgboost and cgboost6 have low average predicting error in each year and each index, both MAPE and Theil U, and predicting result has higher linear correlation with actual price than based experiment. This result proved that proposed model can introduce a more accurate prediction than deep learning. Besides, the result of CGBoost6 is much better than that CGBoost, it is also answer the second question.

%Proposed method is also better than WSAEs-LSTM, which was presented in \cite{bao2017deep}. Our predicting result has higher linear correlation, lower MAPE and Theil U with actual price than WSAEs-LSTM. So it illustrates that GBoost with 1D CNN is more suitable than WSAEs-LSTM in stock time series prediction. In the prediction of CSI 300 index, for example, the mean of MAPE and Theil U of C1D-CGBoost reach 0.014 and 0.009, respectively, which is much less than the result of the WSAEs-LSTM. Besides, the linear correlation has an mean of 0.969, which is also the higher than the model in \cite{bao2017deep}. In fact, C1D-CGBoost outperforms the WSAEs-LSTM not only on mean but also in each year. This is enough to show that our model is more robust than WSAEs-LSTM.

\begin{figure*}
\begin{center}
\includegraphics[width=123mm]{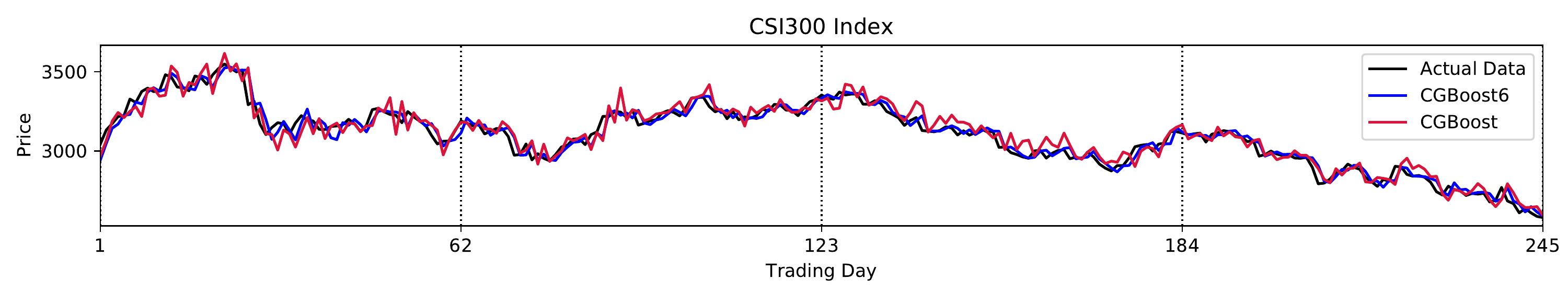}
\includegraphics[width=123mm]{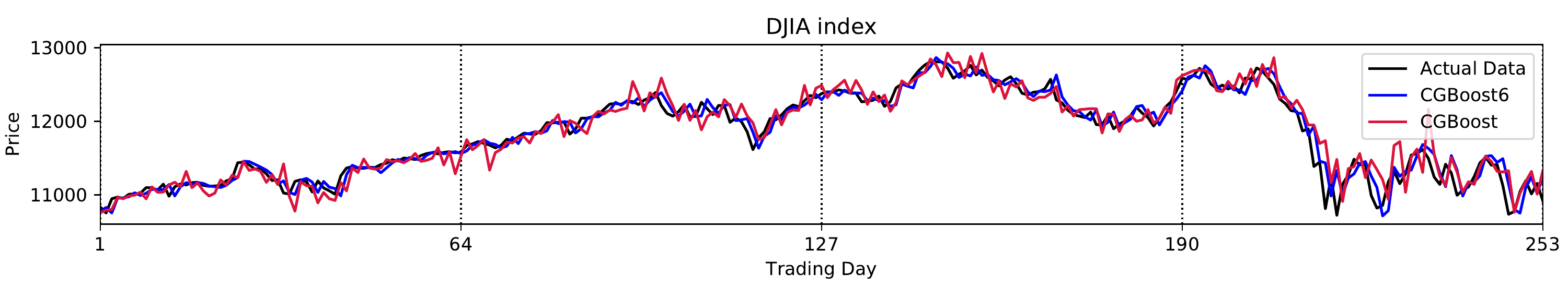}
\includegraphics[width=123mm]{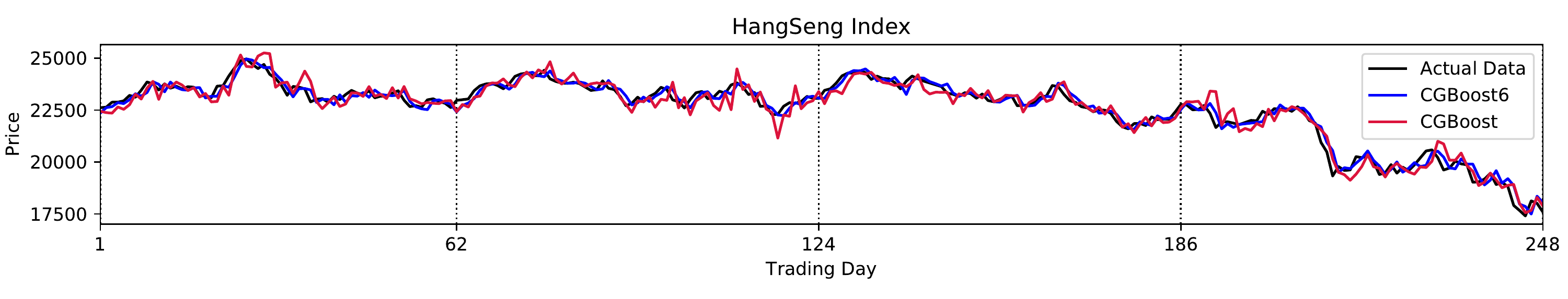}
\includegraphics[width=123mm]{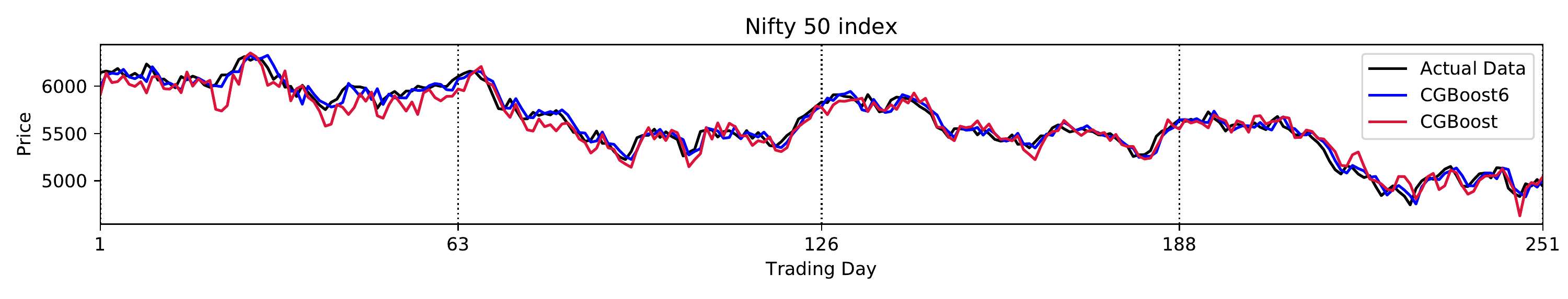}
\includegraphics[width=123mm]{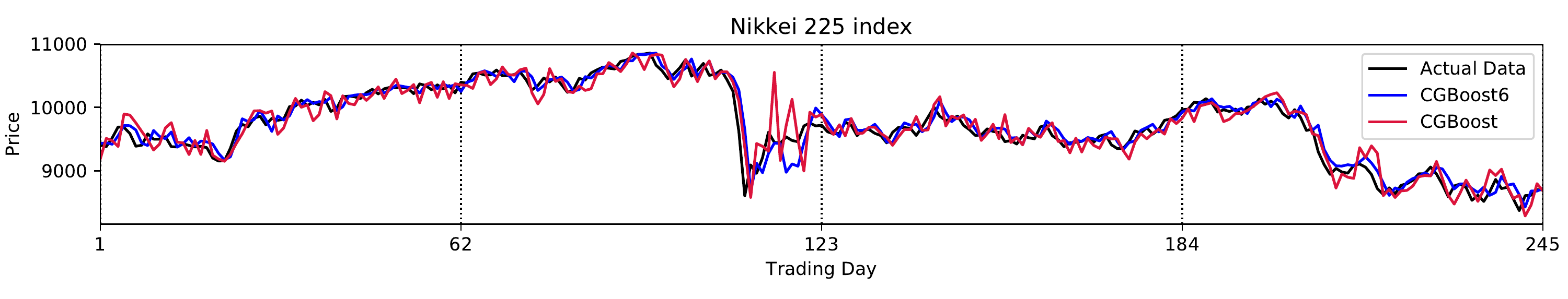}
\includegraphics[width=123mm]{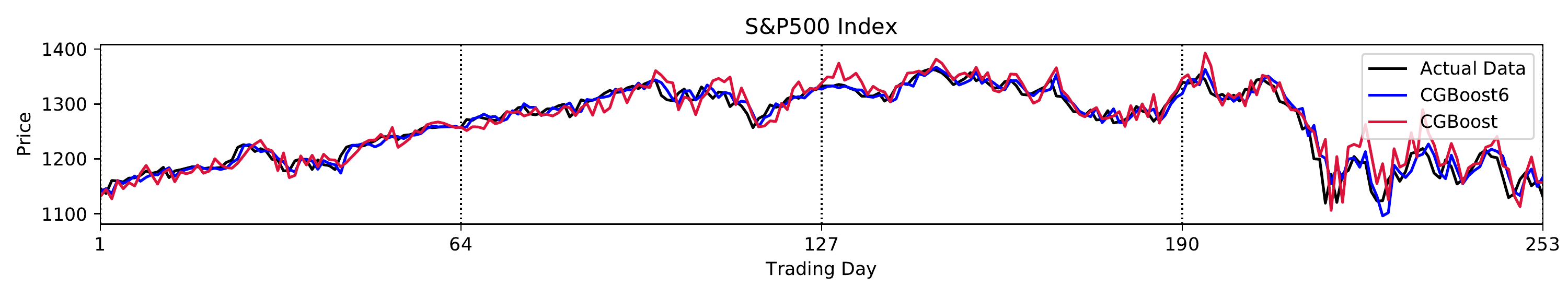}
\end{center}
   \caption{ Shows the actual curves and predicted curves from the our methods for six stock index from 2010.10.01 to 2011.09.30. }
\label{year1}
\end{figure*}

Fig. \ref{year1} shows an example of the Year 1 predicted price from proposed model and the corresponding actual price. CGBoost6 is closer to the actual stock price time series than CGBoost and has lower volatility.

%------------------------------------------------------------------------
\section{Conclusion and Future Work}\label{conclusion}
In this paper we built a new predicting framework to forecast the next day stock price of six stock indices from the financial markets from different country. The process for building this predicting framework is: First, clipping the high value and normalizing the technical index and other feature. Second, using 1D resnet SAEs to denoise and reduce the dimension of features. Last, CGBoost was used to predict the next day price, this is a supervised manner. Our input features include the daily technical indicators, OHLC variables and macroeconomic variables. The main contribution of this paper is attempting to combine 1D resnet with GBoost, a kind of ensemble learning method in stock predict, and prove its performance. Besides, we successfully trains one model on different market and obtain a better prediction on the overall test set.

Future work could focus on increasing the diversity of based estimators. We may try to replace same construction of basic 1D CNNs with several different constructions, in order to improve the performance of CGBoost. Another interesting direction is to use CGBoost in other fields. CGBoost may be applicable to the sequence data including several time series of different features, such as weather forecast, traffic forecast and etc. CGBoost may be able to get better performance in these field.

\section*{Acknowledgment}
This work was supported by the National Natural Science Foundation of China (No.61673322, 61673326, and 91746103), the Fundamental Research Funds for the Central Universities (No. 20720190142), Natural Science Foundation of Fujian Province of China (No. 2017J01128 and 2017J01129), and the European Union's Horizon 2020 research and innovation programme under the Marie Sklodowska-Curie grant agreement (No. 663830).
\bibliographystyle{spmpsci}
\bibliography{mybibfile}
\end{document}